\documentclass[sigconf]{acmart}

\usepackage{threeparttable}
\usepackage{makecell}
\usepackage{multirow}
\usepackage{diagbox}
\usepackage{graphicx}
\usepackage{booktabs}
\usepackage[section]{placeins}
\AtBeginDocument{%
  \providecommand\BibTeX{{%
    \normalfont B\kern-0.5em{\scshape i\kern-0.25em b}\kern-0.8em\TeX}}}

\setcopyright{acmcopyright}
\copyrightyear{2018}
\acmYear{2018}
\acmDOI{XXXXXXX.XXXXXXX}

\acmConference[Conference acronym 'XX]{Make sure to enter the correct
  conference title from your rights confirmation emai}{June 03--05,
  2018}{Woodstock, NY}
\acmPrice{15.00}
\acmISBN{978-1-4503-XXXX-X/18/06}

\begin{document}

% \title[A New Calibration Method DESC Fused Different Fields with Attention on Online Advertising]{Deep Ensemble Shape Calibration : A New Calibration Method Fused Different Fields with Attention on Online Advertising}

\title[Multi-Field Post-hoc Calibration on Online Advertising]{Deep Ensemble Shape Calibration: Multi-Field Post-hoc Calibration in Online Advertising}
\author{Shuai Yang}
%\orcid{}
\affiliation{%
  \institution{Shopee Discovery Ads}
  \city{Beijing}
  \country{China}
  }
%\email{xxx@shopee.com}
\author{Hao Yang}
%\orcid{}
\affiliation{%
  \institution{Shopee Discovery Ads}
  \city{Beijing}
  \country{China}
  }
%\email{xxx@shopee.com}
\author{Zhuang Zou}
%\orcid{}
\affiliation{%
  \institution{Shopee Discovery Ads}
  \city{Beijing}
  \country{China}
  }
%\email{xxx@shopee.com}
\author{Linhe Xu}
%\orcid{}
\affiliation{%
  \institution{Shopee Discovery Ads}
  \city{Beijing}
  \country{China}
  }
%\email{xxx@shopee.com}
\author{Shuo Yuan}
%\orcid{}
\affiliation{%
  \institution{Shopee Discovery Ads}
  \city{Beijing}
  \country{China}
  }
%\email{xxx@shopee.com}
\author{Yifan Zeng}
%\orcid{}
\affiliation{%
  \institution{Shopee Discovery Ads}
  \city{Beijing}
  \country{China}
  }
%\email{xxx@shopee.com}


%% zouzhuang @ apple
%% 1.第一句的句式和FAC的太像了
%% 2.Multi-Field shape calibration and validation 是不是应该提到？
%% 3.attention没有提到

%% zouzhuang @ lucas

\begin{abstract}

% In the e-commerce advertising scenario, estimating the true probabilities (known as a calibrated estimate) on Click-Through Rate (CTR) and Conversion Rate (CVR) is critical. Previous research has introduced numerous solutions for addressing the calibration problem. These methods typically involve the training of calibrators using a validation set and subsequently applying these calibrators to correct the original estimated values during online inference. However, what sets e-commerce advertising scenarios is the challenge of multi-field calibration. Multi-field calibration can be subdivided into two distinct sub-problems: \textbf{value} calibration and \textbf{shape} calibration. \textbf{Value} calibration is defined as no over- or under-estimation for each value under concerned fields (such as the average of pCTR should equal to the CTR for value "women’s shoes" in the field "category", in the advertising context, the number of fields can range from dozens to even hundreds). \textbf{Shape} calibration is defined as no over- or under-estimation for each subset of the pCTR within the specified range. In order to achieve \textbf{shape} calibration and \textbf{value} calibration, it is necessary to have a strong data utilization ability. Because the quantity of pCTR specified range for single field-value (such as user ID and item ID) sample is relative small, which makes the calibrator more difficult to train. However the existing methods cannot simultaneously fulfill both value calibration and shape calibration.  

In the e-commerce advertising scenario, estimating the true probabilities (known as a calibrated estimate) on Click-Through Rate (CTR) and Conversion Rate (CVR) is critical. Previous research has introduced numerous solutions for addressing the calibration problem. These methods typically involve the training of calibrators using a validation set and subsequently applying these calibrators to correct the original estimated values during online inference. 

% However, what sets e-commerce advertising scenarios is the challenge of multi-field calibration. From the perspective of business understanding multi-field, calibration can be subdivided into two distinct sub-problems: \textbf{value} calibration and \textbf{shape} calibration. \textbf{Value} calibration is defined as no over- or under-estimation for each value under concerned fields (such as the average of pCTR should equal to the CTR for value "women’s shoes" in the field "category"). \textbf{Shape} calibration is defined as no over- or under-estimation for each subset of the pCTR within the specified range. In order to achieve \textbf{shape} calibration and \textbf{value} calibration, it is necessary to have a strong data utilization ability. Because the quantity of pCTR specified range for single field-value (such as user ID and item ID) sample is relative small, which makes the calibrator more difficult to train. 

However, what sets e-commerce advertising scenarios is the challenge of multi-field calibration. Multi-field calibration requires achieving calibration in each field. In order to achieve multi-field calibration, it is necessary to have a strong data utilization ability. Because the quantity of pCTR specified range for single field-value (such as user ID and item ID) sample is relatively small, which makes the calibrator more difficult to train. However, existing methods have difficulty effectively addressing these issues.

% 我又想到一点：ValueCalibration 可以被认为在消除预估的不公平现象
% Because the quantity of pCTR specified range for single field-value (such as user ID and item ID) sample is relative small, which makes the calibrator more difficult to train. 

% Previous research has introduced numerous solutions for addressing the calibration problem. These methods typically involve the training of calibrators using a validation set and subsequently applying these calibrators to correct the original estimated values during online inference. However, what sets e-commerce advertising scenarios is the challenge of multi-field calibration.  
%在多特征校准的要求下，任何一条流量预估都是精准的：不管pctr在什么区间，以及从哪一个特征取值的角度，预估得分都是精准的，这对数据利用率提出了更高的要求，而现有方法无法满足这些条件%

%传统的校准方法在多特征校准上会遇到如下问题：有些方法无法保证pctr在每个区间都是准的；有些方法无法保证所有特征值整体是准的%

% To solve these problems, we propose a new method named \textbf{D}eep \textbf{E}nsemble \textbf{S}hape \textbf{C}alibration (DESC) \footnote{The code is at \href{https://github.com/HaoYang0123/DESC}{https://github.com/HaoYang0123/DESC}}. We introduce innovative basis calibration functions, which enhance both function expression capabilities and data utilization by combining these basis calibration functions. A significant advancement lies in the development of an allocator capable of allocating the most suitable shape calibrators to different estimation error distributions within diverse fields and values. Finally, we achieve significant improvements in both public data and industrial data, and on online experiments we get +2.5\% of CVR and +4.0\% of GMV (Gross Merchandise Volume).  

To solve these problems, we propose a new method named \textbf{D}eep \textbf{E}nsemble \textbf{S}hape \textbf{C}alibration (DESC).
In terms of business understanding and interpretability, we decompose multi-field calibration into \textbf{value} calibration and \textbf{shape} calibration. We introduce innovative basis calibration functions, which enhance both function expression capabilities and data utilization by combining these basis calibration functions. A significant advancement lies in the development of an allocator capable of allocating the most suitable calibrators to different estimation error distributions within diverse fields and values. We achieve significant improvements in both public and industrial datasets. In online experiments, we observe a +2.5\% increase in CVR and +4.0\% in GMV. (Gross Merchandise Volume). Our code is now available at:
https://github.com/HaoYang0123/DESC.

% To solve these problems, we propose a new method named \textbf{D}eep \textbf{E}nsemble \textbf{S}hape \textbf{C}alibration (DESC) \footnote{The code is at \href{https://github.com/HaoYang0123/DESC}{https://github.com/HaoYang0123/DESC}}. We introduce innovative basis calibration functions, which enhance both function expression capabilities and data utilization by combining these basis calibration functions. A significant advancement lies in the development of an allocator capable of allocating the most suitable calibrators to different estimation error distributions within diverse fields and values. Finally, we achieve significant improvements in both public data and industrial data, and on online experiments we get +2.5\% of CVR and +4.0\% of GMV (Gross Merchandise Volume).  

\end{abstract}



\maketitle
% \vspace{-15pt}
\section{INTRODUCTION}
\begin{figure}
\setlength{\abovecaptionskip}{0.0cm}
\setlength{\belowcaptionskip}{-0.2cm}
    \centering
    \includegraphics[width=1.0\linewidth]{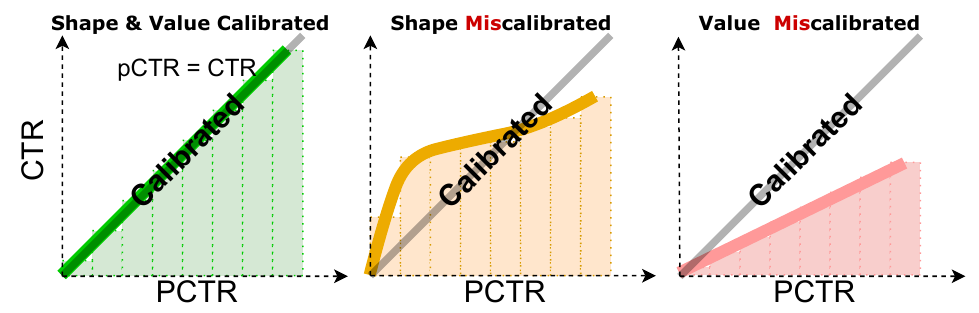}
    \caption{Examples to show the shape miscalibration and value miscalibration}
    \label{shape_and_value}
    \vspace{-10pt}
\end{figure}
% \begin{figure}
% \setlength{\abovecaptionskip}{0.0cm}
% \setlength{\belowcaptionskip}{-0.2cm}
%     \centering
%     \includegraphics[width=1.0\linewidth]{shape_and_value_miscalib_2.pdf}
%     \caption{Example to show shape miscalibration and value miscalibration from the distributions of pCTR and CTR}
%     \label{fig:enter-label}
%         % \vspace{-15pt}
% \end{figure}

Estimating CTR and CVR is a crucial technology in e-commerce advertising domains \cite{cheng2016wide,shan2016deep,zhai2016deepintent,zhu2017optimized,huang2019fibinet,wang2021masknet,du2021exploration,davidson2010youtube,guo2017deepfm,wang2018billion,bian2022can,nguyen2023lightsage,yang2023practice,deng2021calibrating}. Accurate prediction of CTR (pCTR) and CVR (pCVR) is essential, as it necessitates precision not only in ranking but also in absolute values.

However, recent studies \cite{bella2010calibration,guo2017calibration,korb1999calibration,ovadia2019can,feldman2023calibrated} have revealed that numerous established machine learning techniques, particularly deep learning methods extensively applied in the fields of e-commerce advertising, often yield inadequately calibrated probability predictions.

\begin{figure}
\setlength{\abovecaptionskip}{0.0cm}
\setlength{\belowcaptionskip}{-0.2cm}
    \centering
    \includegraphics[width=0.7\linewidth]{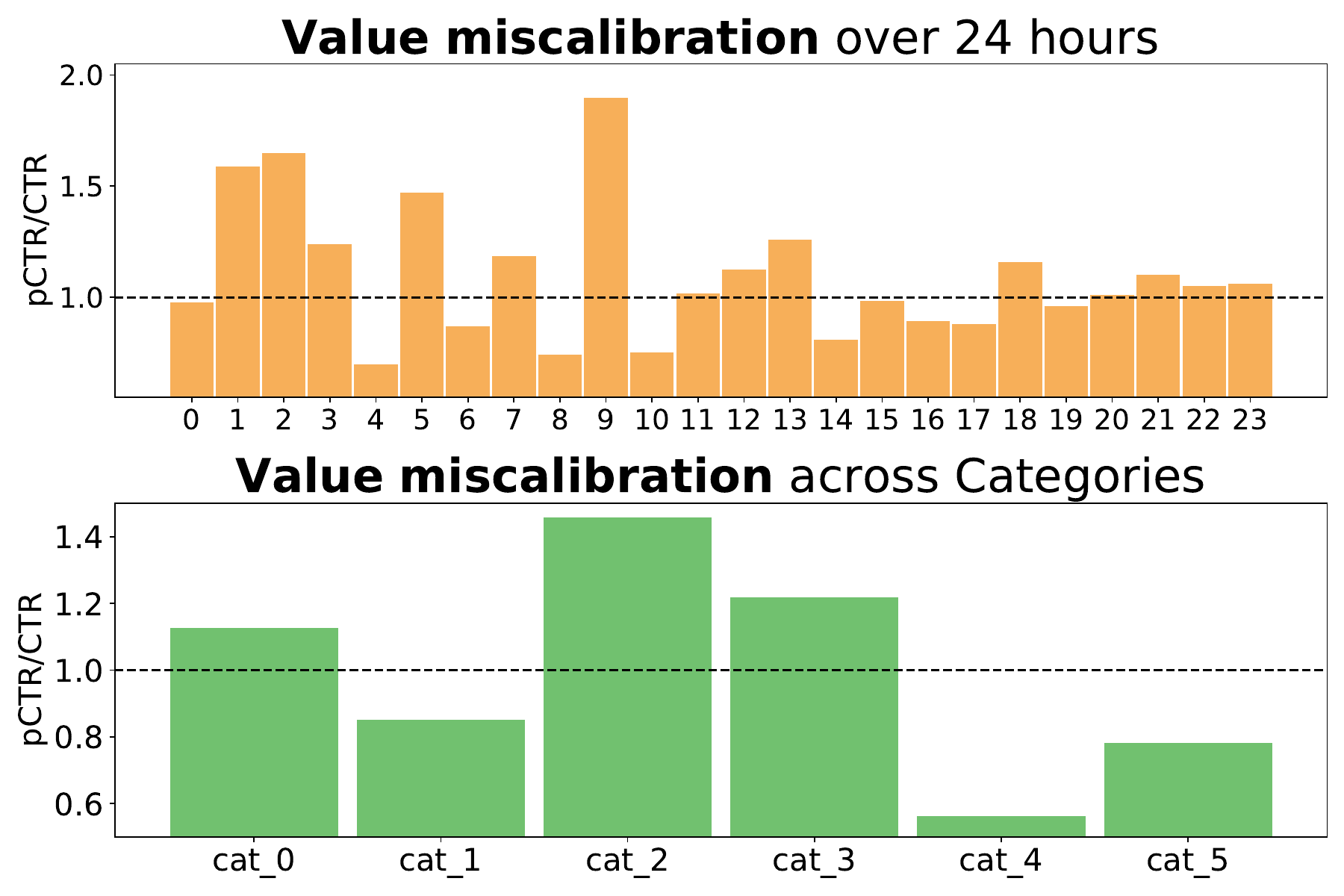}
    \caption{Examples to show significant variations in value miscalibration among different fields and values.}
    \label{fig:case=dif-fv}
    \vspace{-10pt}
\end{figure}

\begin{figure}
\setlength{\abovecaptionskip}{0.0cm}
\setlength{\belowcaptionskip}{0.0cm}
    \centering
    \includegraphics[width=0.8\linewidth]{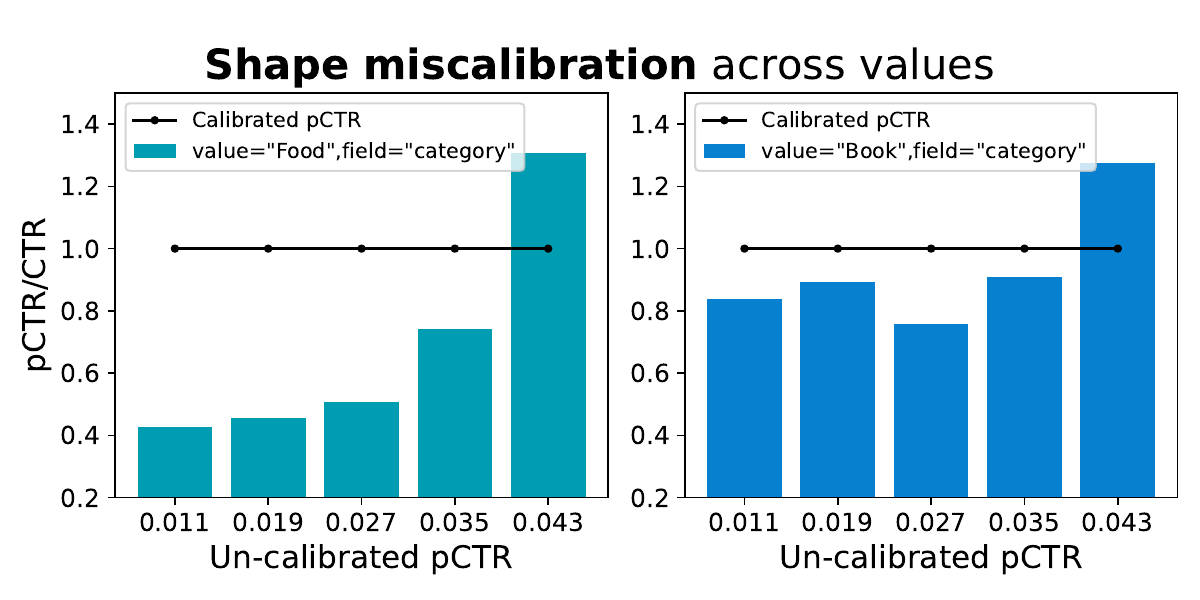}
    \caption{Examples to show significant variations in shape miscalibration among different values within the same field.} %Samples sharing the same value within a given field are sorted by their non-calibrated pCTR values (x-axis), subsequently divided into equal-frequency bins, with the real CTR values (y-axis) computed for each bin.}
    \label{fig-example}
    % \vspace{-15pt}
    \vspace{-5pt}
\end{figure}

In previous research, there are many solutions to solve the calibration problem. These methods learn calibrator through the validation set, and then use the calibrator to correct the original estimated values in the online inference stage. These methods can be categorized into parameter methods (including Platt Scaling \cite{platt1999probabilistic}, Temperature scaling \cite{guo2017calibration}, Beta calibration \cite{kull2017beta}, Gamma calibration \cite{kweon2022obtaining} and Dirichlet calibration \cite{kull2019beyond}), non-parameter methods (such as Histogram Binnning \cite{zadrozny2001obtaining} and Isotonic Regression \cite{zadrozny2002transforming}) and hybrid methods. Hybrid methods encompass both non-field-aware \cite{kumar2019verified, jiang2011smooth, zhang2020mix} and field-aware approaches \cite{pan2020field, wei2022posterior, huang2022mbct}.

In the advertising scenarios, there are multiple fields (such as user group and item category), and if predictions of CTR (pCTR) or CVR (pCVR) are not calibrated in certain fields, such as item category, it can lead to a negative impact on the earnings of some advertisers. 
Simultaneously, in the e-commerce scenario, there is a particularly large number of fields to consider. This necessitates the calibration of each field, referred to as \textbf{multi-field calibration}. To achieve multi-field calibration, it is necessary to have a strong data utilization ability. Because the samples of the specified pCTR range for a single field value (such as user ID and item ID) are relatively small, which makes the calibrator more difficult to train.

In terms of business understanding and interpretability, we decompose multi-field calibration into \textbf{value} calibration and \textbf{shape} calibration. Figure \ref{shape_and_value} shows \textbf{shape} miscalibration and \textbf{value} miscalibration. \textbf{Value} calibration is defined as no over- or under-estimation each value under concerned fields
(such as average of pCTR should equal to the CTR for value "women’s shoes" in the field "category", in the e-commerce advertising context, the number of fields can range from dozens to even hundreds). From the advertising perspective, \textbf{value} calibration ensures that the ECPM (Effective Cost Per Mille) and GMV (Gross Merchandise Value) of different items are not over- or under estimated. Figure \ref{fig:case=dif-fv} illustrates the significant inconsistency of over- and under-estimation across different values. 
\textbf{Shape} calibration is defined as no over- or under-estimation for each subset of the pCTR within the specified range. Explaining from the advertising perspective, \textbf{shape} calibration ensures that some already popular items are not excessively exposed or suppressed. Take Figure \ref{fig-example} as an example, for the field "Category" with values "Food" and "Book", the calibrated pCTR/CTR is 1 in any pCTR interval, whereas the distribution (shape) of the uncalibrated pCTR/CTR will be different. However, the existing methods cannot simultaneously fulfill both value calibration and shape calibration.

For parameter methods, non-parameter methods and non-field-aware methods, they involve the training of a single calibration function to address the issue of over- or under-estimation for each subset of the pCTR globally but overlook the biases across different fields.

For field-aware methods, they have different emphases, and cannot simultaneously fulfill both value calibration and shape calibration. NeuralCalib/Field-aware Calibration (FAC) \cite{pan2020field} places its focus on modeling estimation value biases across various field values, rather than addressing shape miscalibrations. Multiple Boosting Calibration Tree (MBCT) \cite{huang2022mbct} effectively enhances the binning of field values, giving less emphasis to shape miscalibrations. AdaCalib (Ada) \cite{wei2022posterior} can only model shape calibration and value calibration under the condition of a single field.

An analysis of the aforementioned field-aware methods reveals several key observations. Firstly, within a single field, there exists multiple values. In some cases, the allocation of samples to each field value may be limited, particularly in scenarios involving user ID, item ID and certain sparse cross-features. Consequently, training a calibrator under such limited samples can be challenging. Additionally, field-aware methods often employ a binning strategy, followed by the generation of calibration parameters based on information from two adjacent bins. This approach can lead to a sample isolation issue wherein the parameters of a bin are exclusively influenced by the samples within that bin and its neighboring bins, with no updates from samples in other bins. As a result, these methods can result in suboptimal data utilization.

To solve these above problems, we propose a new method named \textbf{D}eep \textbf{E}nsemble \textbf{S}hape \textbf{C}alibration (\textbf{DESC}). There are four contributions in our work:

\vspace{-5pt}
\begin{itemize}
    \item We redefine the multi-field calibration: perform shape calibration and value calibration at the same time.
    \item We propose the novel basis calibration functions, which can simultaneously improve function expression ability and data utilization through the combination of basis calibration functions.
    \item We make a breakthrough in putting forward an allocator that can allocate the most suitable shape calibrators for different estimation error distributions on various fields and values.
    \item Our proposed DESC method outperforms other methods significantly across both calibration and non-calibration metrics, as demonstrated on two public datasets and one industrial dataset. Additionally, in online experiments, we observe a notable increase of +2.5\% in CVR and +4.0\% in GMV.
\end{itemize}

\vspace{-10pt}
\section{RELATED WORKS}

With the growing emphasis on improving the reliability and accuracy of machine learning models, several research approaches have emerged. Numerous calibration methods focus on acquiring a mapping function to convert predicted probabilities into observed posterior probabilities, referred to as post-hoc calibration. These studies can be broadly classified into three categories: non-parameter, parameter, and hybrid methods.

\vspace{-10pt}
\subsection{Non-parametric Methods} %罗列各无参方法: Hisgram binning, Isotonic Regression, BBQ

%%lucasyang:Non-parametric methods do not have distribution assumptions 可以先介绍下为何叫非参，参照MBCT%
% 列举非参方法的问题：数据利用率，在边界点性能问题
% 调换下非参和参数方法的顺序

% Non-parametric methods make no assumptions about the distribution of estimates. Non-parametric methods include Histogram Binning \cite{zadrozny2001obtaining} and Isotonic Regression \cite{zadrozny2002transforming}. Histogram Binning sorts the estimated values, then divides them into bins at equal frequencies or intervals, and uses the posterior CTR of each bin as a calibration parameter. On this basis, Isotonic Regression adds the requirement of order preservation: if the CTR of the later bin is lower than the CTR of the previous bin, then merge the two buckets to achieve the effect of order preservation. In addition, this method will lead to instability of the bucket boundary value.

Non-parametric methods do not rely on any assumptions regarding the distribution of estimates. Examples of non-parametric methods encompass Histogram Binning (HB)\cite{zadrozny2001obtaining} and Isotonic Regression (IR) \cite{zadrozny2002transforming}. HB involves sorting the estimated values and then dividing them into bins of equal frequencies or intervals. Building upon this, IR introduces the additional constraint of order preservation. These approach can introduce instability in the bin boundary values.

\vspace{-10pt}
\subsection{Parametric Methods} %罗列有参方法: Plat, TS, Gamma/Beta/Dirichlet
% Parameter methods typically assume that the probabilities follow a certain distribution, so the mapping function can be derived on specific conditions. Platt Scaling \cite{platt1999probabilistic} is widely used in binary classification calibration, which adopts Guassian assumption with the same variance of the positive and negative class \cite{kweon2022obtaining}. Temperature scaling \cite{guo2017calibration} extends to multi-class task. Beta calibration \cite{kull2017beta} introduces the Beta distribution into the predicted probabilities. Similarly, Gamma calibration \cite{kweon2022obtaining} and Dirichlet calibration \cite{kull2019beyond} use their probability distribution to calibrate. Parameter methods severely depend on the strong distribution assumption. Consequently, the performance may be sub-optimal if the assumption does not hold in practice. 

Parameter methods often rely on specific probability distribution assumptions for deriving mapping functions. Platt Scaling \cite{platt1999probabilistic}, extensively employed in binary classification calibration, assumes a Gaussian distribution with equal variances for both positive and negative classes \cite{kweon2022obtaining}. For multi-class tasks, Temperature Scaling \cite{guo2017calibration} extends this approach. Beta calibration \cite{kull2017beta}, Gamma calibration \cite{kweon2022obtaining} and Dirichlet calibration \cite{kull2019beyond} rely on their respective probability distributions for calibration. Parameter methods are highly reliant on the strong distribution assumption, which can lead to suboptimal performance if the assumption does not hold in practical scenarios.

\vspace{-5pt}
\subsection{Hybrid Methods} 
% Hybrid methods combine non-parametric and parametric methods. Depending on whether performed on the field-level, we further divide hybrids into two groups: non-field-aware and field-aware methods.

Hybrid methods integrate non-parametric and parametric approaches. Based on whether they are applied at the field-level, we further categorize hybrids into two groups: non-field-aware and field-aware methods.

\vspace{-5pt}

\subsubsection{Non-field-aware Methods} 
Non-field-aware methods encompass three notable approaches: Scaling-binning \cite{kumar2019verified}, Smooth Isotonic Regression (SIR) \cite{jiang2011smooth}, and Ensemble Temperature Scaling (ETS) \cite{zhang2020mix}. Scaling-binning combines the techniques of Platt Scaling and Histogram Binning. SIR employs linear interpolation based on isotonic regression, while ETS combines multiple temperature scalings for calibration. Notably, all of these methods utilize raw predicted scores as input without taking field information into account.

\vspace{-5pt}
\subsubsection{Field-aware Methods} 
Field-aware methods integrate supplementary features to formulate calibration functions. NeuralCalib (FAC) \cite{pan2020field} introduces an auxiliary module but it does not address the challenge of shape miscalibration variance among different field values. AdaCalib (Ada) \cite{wei2022posterior} learns an isotonic function based on posterior statistics and selects the most appropriate bin number for a single field value. However, AdaCalib does not encompass all fields. MBCT \cite{huang2022mbct} employs trees to uncover more effective calibration across fields. However, calibration trees are unable to handle sparse fields, such as user ID and item ID.

% As depicted in Table \ref{tab-require}, none of the aforementioned methods can simultaneously tackle all three of these challenges
\vspace{-5pt}
\section{CALIBRATION PROBLEM FORMULATION}
% 定义问题：
% 第一段：数据分三个：1）Train；2）Dev；3）Test。步骤1: Train数据集用于训练无校准模型A；步骤2: 使用A模型在Dev（或者Dev+Train）数据上得到点击率概率，然后用于训练校准模型B；步骤3:最后在Test数据集上评测校准模型B的效果
In online advertising systems, using CTR as an illustration (CVR follows the same principles), we have trained a neural predictor, denoted as $f_{uncalib}$, on a training dataset $\mathcal{D}_{train}$. This dataset includes all field values as inputs ($x$) and click responses ($y$), where $y=0$ represents non-click events and $y=1$ signifies click events. The neural predictor is capable of forecasting the likelihood of a click using the following formula.
\begin{small}
\begin{equation}
\label{eq-uncalib}
    \hat{p}_{uncalib} = f_{uncalib}(x)
\end{equation}
\end{small}

To address the under- and over-estimation issues associated with the predicted scores generated by $f_{uncalib}$, we must train an additional calibrator, denoted as $f_{calib}$, while considering $n$ distinct fields ($Z_1$, $Z_2$, ..., $Z_n$) in a validation dataset. Our objective is for $f_{calib}$ to predict the conditional expectation $\mathbf{E}[y|x]$. Subsequently, the theoretical calibration error of $f_{calib}$ with respect to the ground-truth under the $l_p$-norm is defined as:

\vspace{-10pt}
\begin{small}
\begin{equation}
\label{eq-cal-tce}
TCE_p(f_{calib}) = (\mathbf{E}_x[|\mathbf{E}[y|x] - f_{calib}(x)|^p])^{\frac{1}{p}} .
\end{equation}
\end{small}

The calibrator $f_{calib}$ is considered to be perfectly calibrated when the theoretical calibration error $TCE_p(f_{calib})$ is zero. However, in fact perfect calibration is impossible in practice. Only approximate and asymptotic grouped calibrations are possible for finite and specific partitions of samples \cite{gupta2020distribution}. To test the performance of different calibrators, we will explain some related calibration metrics in EXPERIMENTS section.

% \vspace{-5pt}
% \section{MULTI-FIELD CALIBRATION PROBLEM FORMULATION}

% \begin{equation}
%     ECE_k = \frac{1}{m}\sum_{i=1}^{m}ABS(\frac{pCTR_m}{CTR_m} - 1)
% \end{equation}

% \begin{equation}
%     CTR_m = \frac{\sum(Click)}{\sum(Impression)}
% \end{equation}

% %多特征校准误差被定义为%

% \begin{equation}
%     MF\text{-}ECE = \frac{1}{n} \sum_i F\text{-}ECE_{i} 
% \end{equation}
% \begin{equation}
%     MF\text{-}ECE = \frac{1}{fN} \sum_{fn=1}^{fN} ( \frac{1}{vN} \sum_{vn=1}^{vN} (\frac{1}{bN} \sum_{bn=1}^{bN} abs(\frac{pCTR_{fn,vn,bn}}{CTR_{fn,vn,bn}}-1))) 
% \end{equation}

\vspace{-5pt}
\section{METHODS}

\begin{figure*}
\setlength{\abovecaptionskip}{0cm}
\setlength{\belowcaptionskip}{0cm}
    \centering
    \includegraphics[width=0.85\textwidth]{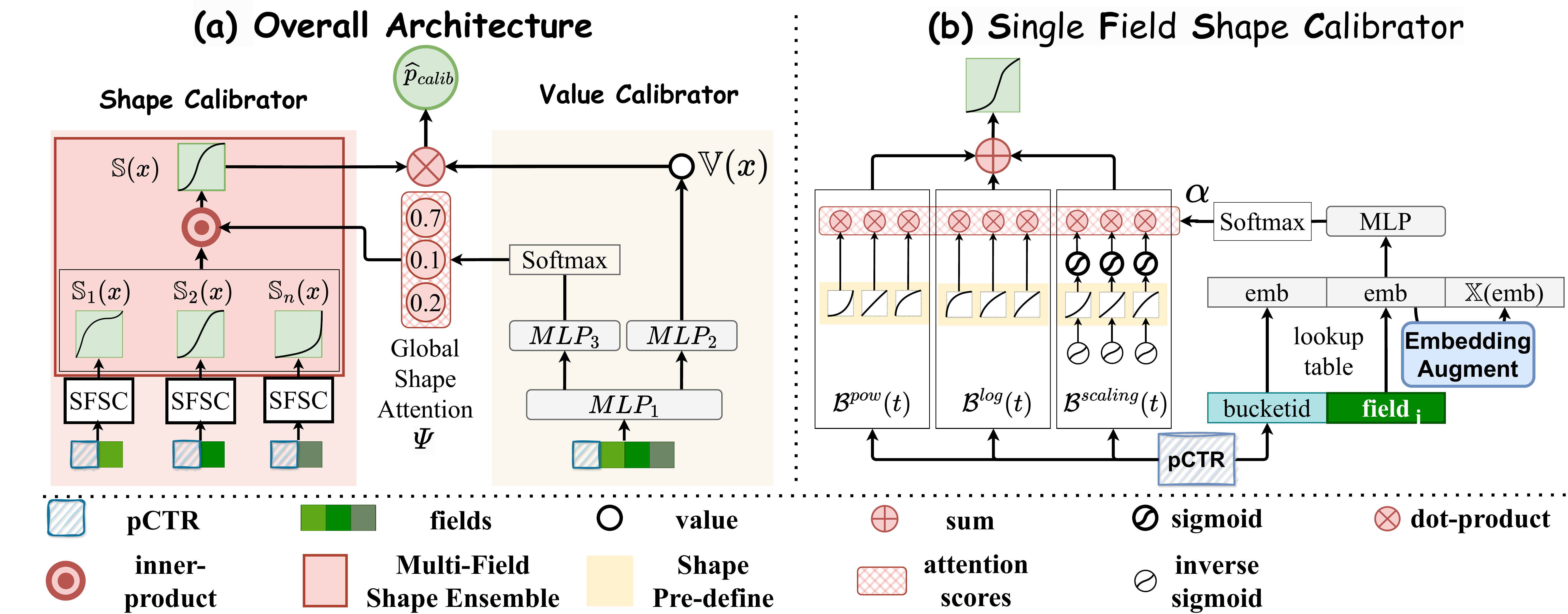}
    \caption{(a) Overall architecture of DESC, its input includes the non-calibrated pCTR and the original fields, and the final output is the calibrated pCTR. It is end-to-end trainable. (b) Single Field Shape Calibrator takes one field and the non-calibrated pCTR as inputs and outputs the calibrated shape score for this field.}
    \label{fig:enter-label-overall}
    \vspace{-10pt}
\end{figure*}
% 我们将校准的问题定义为两部分：shape Calibration 以及 Value Calibration，然后最终的校准 = Shape * Value

Under the requirement of multi-field calibration, both \textbf{shape} calibration and \textbf{value} calibration need to be performed simultaneously. Therefore, within the entire DESC architecture, we have designed separate modules, namely Shape Calibrator and Value Calibrator (Figure \ref{fig:enter-label-overall}), to achieve \textbf{shape} calibration and \textbf{value} calibration. The final calibrated score is the product of these two parts:
\begin{small}
\begin{equation}
    \label{eq-final-score}
    \hat{p}_{calib} = \mathbb{S}(x) \cdot \mathbb{V}(x) \ ,\ x = \{Z,\hat{p}_{uncalib} \}
\end{equation}
\end{small}

% \begin{small}
% \begin{equation}
%     \label{eq-final-score}
%     \hat{p}_{calib} = \mathbb{S}(x,\hat{p}_{uncalib} ) \cdot \mathbb{V}(x,\hat{p}_{uncalib} ) .
% \end{equation}
% \end{small}

The input $x$ consists of all $n$ field values $Z$ and non-calibrated scores $\hat{p}_{uncalib}$.
$\mathbb{S}(x)$ and $\mathbb{V}(x)$ refer to the shape calibrated score (the output of Shape Calibrator) and value calibrated score (the output of Value Calibrator), respectively. 
The training loss function is the negative log-likelihood function. 
\begin{small}
\begin{equation} 
 \mathcal{L}(x, \; y) = -\frac{1}{|\mathcal{D}|}(y \cdot log \hat{p}_{calib} + (1-y) \cdot log(1-\hat{p}_{calib}))
\end{equation}
\end{small}

In Section 4.1, we present the Shape Calibrator module, which is responsible for achieving shape calibration.
In Section 4.2, we discuss the Value Calibrator module, which is designed to accomplish value calibration.
In Section 4.3, we elaborate on the deployment of DESC in an online setting, outlining the necessary procedures and considerations.

\vspace{-5pt}
\subsection{Shape Calibrator}
%%
%Shape Calibration的作用，在给定特征的条件下，对于pctr的所有区间，都不会存在高估或者低估的情况。
For multi-field calibration, the goal of Shape Calibrator is to ensure that: given input features $x$, we need to reduce the problems of over- and under-estimation across all intervals of pCTR (shape miscalibration).
% PCTR是不是应该换一个说法？

In section 4.1.1, we pre-define a variety of basis functions to accommodate different shape requirements. Section 4.1.2 deals with the allocation of appropriate shape functions given specific field conditions. This section offers separate discussions on how shape allocation is managed for regular fields and sparse fields.

In section 4.1.3, when dealing with multiple fields, there may be conflicts in terms of the influence of different fields on calibration. Thus, we introduce a multi-field fusion mechanism named Multi-Field Shape Ensemble.

Collectively, sections 4.1.1 and 4.1.2 are referred to as the Single Field Shape Calibrator (SFSC), as depicted in Figure
\ref{fig:enter-label-overall}b.

\subsubsection{Shape Pre-Define}

%在Shape Pre-Define，我们预定义了一些基函数，包括幂函数，Scaling函数，log函数等，其中这些基函数的参数可以学习得到。
In this section, we pre-define some basis calibration functions. Then, we merge the basis calibration functions into shape functions to enhance their expressive ability. 
 % Afterward, we concatenate the different parameters from basis functions to learn combined shapes.

For the variable $t$, $t$ is between 0 and 1, basis calibration function $\mathcal{B}(t)$ satisfies the following conditions:

{\bfseries1.} The function is monotonically non-decreasing and continuous in range $(0,1)$.

{\bfseries2.} When $t$ approaches $0^{+}$, the limit of $\mathcal{B}(t)$ is 0, and when $t$ approaches $1^{-}$, the limit of $\mathcal{B}(t)$ is 1 (Equation \ref{eq-lim}). 

\begin{small}
% \begin{equation}
% \label{eq-mono}
% % \frac{d\mathcal{B}(x)}{dx} >= 0 ,\ where\  x \in (0,1)
% \mathcal{B}(x_1) <= \mathcal{B}(x_2) ,where\ x_1 < x_2 \ and\  x \in (0,1)
% \end{equation}
\begin{equation}
\label{eq-lim}
\lim_{t \to 0^{+}} \mathcal{B}(t) = 0 \ and\ \lim_{x \to 1^{-}} \mathcal{B}(t) = 1 
\end{equation}
\end{small}
 We pre-define $m$ basis functions consisting of $p$ power functions, $l$ logarithmic functions and $s$ scaling functions (shown in Equation \ref{eq-B-x}). These functions exhibit different shape characteristics, and their shapes vary when the hyper-parameters (\emph{e.g.} the coefficients in these basis calibration functions) are different. 
We choose predefined functions over Multi-Layer Perceptron (MLP) because, through data research, we have found that predefined functions can address the issue of over- or underestimation for each subset of pCTR within the specified range (shape calibration). Additionally, predefined functions have a lower parameter count and superior performance.

% 我们使用pre-define 函数 而不是 多层全连接是因为 basic function 能满足 shape calibration 的需求并且有有更低的参数量和更高的性能。通过对数据的分析我们发现，通过log，exp，scaling 函数的组合可以简单的取得较好的效果。
% $

% TODO Online 部分加1个耗时（5ms）

Compared to segmented linear functions in traditional calibration methods, including SIR, FAC, Ada, using basis calibration functions doesn't require data segmentation. In other words, shape learning can utilize all samples for training, significantly enhancing the data utilization. Basis calibration functions include, but are not limited to $log$, $exp$ and $scaling$. By analyzing the data, we find that the combination of $log$, $exp$ and $scaling$ functions can satisfactorily fulfill our requirements with ease.
\begin{small}
\begin{equation}
\begin{matrix}
\label{eq-B-x}
%\mathcal{B}(x) = CAT(\underbrace{ P_{1,1},...,P_{1,pn} }_{ Power function}  \ \ \ \ \underbrace{ L_{1,1},...,L_{1,ln} }_{ Log function} \ \ \ \ \underbrace{ S_{3,1},...,S_{3,sn} }_{ Scaling function} ) 

% \mathcal{B}(x) = \{\underbrace{ P_{1},...,P_{p} }_{ \mathcal{B}^{power}},  \underbrace{ L_{1},...,L_{l} }_{ \mathcal{B}^{log} },  \underbrace{ S_{1},...,S_{s} }_{ \mathcal{B}^{scaling} } \}
% \end{matrix}
% \end{equation}
% \begin{equation}
% \mathcal{B}^{power}(x) = x^{\mathbf{H}}
% \end{equation}
% \begin{equation}
% \mathcal{B}^{log}(x) = \frac{log_2(1 + \mathbf{V} \cdot x)}{log_2(1+\mathbf{V})}
% \end{equation}
% \begin{equation}
% \mathcal{B}^{scaling}(x)  = \sigma( \sigma^{-1}(x) \cdot \mathbf{A} + \mathbf{B} )\ \ ,
% \end{equation}
% \begin{equation}
% where\ \sigma(x) = \frac{1}{1+exp(-x)}\ and\ \sigma^{-1}(x) = log(\frac{x}{1-x})
% \end{equation}
\mathcal{B}(t) = \{\underbrace{ \mathcal{B}_{1}^{power},...,\mathcal{B}_{p}^{power} }_{ p\ power\ functions },  \underbrace{ \mathcal{B}_{1}^{log},...,\mathcal{B}_{l}^{log} }_{ l\ log\ functions },  \underbrace{ \mathcal{B}_{1}^{scaling},...,\mathcal{B}_{s}^{scaling} }_{ s\ scaling\ functions } \}
\end{matrix}
\end{equation}
\begin{equation}
\mathcal{B}_i^{power}(t; h_i) = x^{h_i}
\end{equation}
\begin{equation}
\mathcal{B}_i^{log}(t; v_i) = \frac{log(1 + v_i \cdot t)}{log(1+v_i)}
\end{equation}
\begin{equation}
\mathcal{B}_i^{scaling}(t; a_i)  = \sigma( \sigma^{-1}(t) \cdot a_i)\ \ ,
\end{equation}
\begin{equation}
where\ \sigma(t) = \frac{1}{1+exp(-t)}\ and\ \sigma^{-1}(t) = log(\frac{t}{1-t})
\end{equation}
\end{small}

$h_i, v_i$ and $a_i$ (${h_i,v_i,a_i} \in \mathbf{R}^{+}$) are parameters of basis calibration functions. For each type of basic calibration function, we pre-define these parameters using equally spaced floats (e.g., $[0.1, 0.3, 0.5, 0.7]$), which can be set as trainable.
 
\begin{figure}
        \centering
        \includegraphics[width=1.0\linewidth]{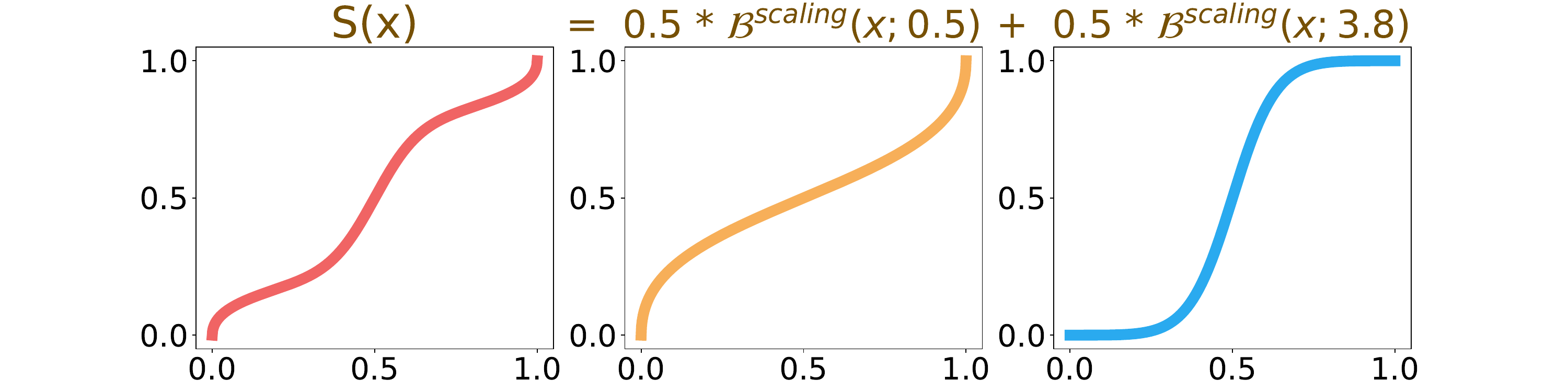}
        \caption{Complex shape can be composed of simple shapes.}
        \label{fig:enter-label-com}
\vspace{-10pt}
\end{figure}

However, $\mathcal{B}(t)$ can only represent simple shapes, and complex shapes can be composed of simple shapes, as shown in Figure \ref{fig:enter-label-com}. Therefore, we need to combine these basis calibration functions through weighted summation to create shape function $\mathbb{S}_i(x)$ capable of representing complex shapes for $i$-th field, as shown in Equation \ref{eq-shape-func}, where $\alpha_i^j$ refers to the attention weight of $j$-th basis function for $i$-th field, which will be explained in next section.

\vspace{-10pt}
\begin{small}
\begin{equation}
\label{eq-shape-func}
\mathbb{S}_i(x) = \sum_{j=1}^{m} { \alpha_i^{j}  \cdot \mathcal{B}_{j}(t) }, \;\; \mathcal{B}_{j}(t) \in \mathcal{B}(t)
\end{equation}
\end{small}
\vspace{-5pt}

% 第二段：评测指标：AUC，logloss，F-RCF，F-ECE@k。特别是解释一下F-ECE@k，相对于F-RCF，F-ECE@k有什么优势

% apple: 上面公式S(x)是通过M个basis calibration function得到，然后又说有M个S(x)，两个M是一个东西吗？

% For the parameters of each Shape Function, we perform $\mathbf{M}$ random initializations, resulting $\mathbf{M}$ individual Shape Functions, which are then concatenated to obtain the Pre-Defined Shape Function.

% \begin{equation}
% \mathcal{\Vec{S}}_{1\times \mathbf{M}} = CAT(\tilde{S}(x, p_1), \tilde{S}(x, p_2),...,\tilde{S}(x, p_M))
% \end{equation}

% \vspace{-15pt}

%% 13中的M和14中的M， 13中的M是不是等于3
\subsubsection{Shape Allocation}

% \vspace{-10pt}
In shape allocation, we allocate suitable shape functions based on feature values. These features include two parts: pCTR bucket feature and original field features. PCTR bucket feature categorizes pCTR into intervals. For example, 0 to 0.001 is bucket 1, and 0.001 to 0.003 is bucket 2. Introducing pCTR bucket feature allows for better shape allocation within different pCTR intervals. The specific steps are as follows:\\
%% step1 + 2不是就是embedding？
% {\bfseries Step1.}  \\
{\bfseries Step1.} We pre-define the embedding size for each feature. Each one-hot feature (including pCTR bucket feature $bucket_{\hat{p}_{uncalib}}$ and original field $Z_i$) is projected into a fixed-size dense embedding, such as $\mathbf{b_{\hat{p}_{uncalib}}}$ and $\mathbf{e_{i}}$. The bucket size of $bucket_{\hat{p}_{uncalib}}$ is assigned as 100. \\
{\bfseries Step2.} We concatenate embedding vectors into a Multi-Layer Perceptron (MLP), followed by a softmax operation, the output $\mathbf{\alpha}_i$ size of softmax is $1\times m$, with $\mathbf{\alpha}_i^1$ to $\mathbf{\alpha}_i^m$, same as the number of pre-defined basis functions.
\begin{small}
\begin{equation}
\label{eq-mlp-ori-wei}
   \mathbf{\alpha}_i = Softmax(MLP(\mathbf{b_{\hat{p}_{uncalib}}}, \; \mathbf{e_{i}}))
\end{equation}
\end{small}
{\bfseries Step3.} Finally, we obtain the output value of the Shape Calibrator $\mathbb{S}_i(x)$ for $Z_{i}$ shown in the previous Equation \ref{eq-shape-func}.

In the shape allocation stage, the expressive ability of embedding directly affects the ability of allocating shapes. For sparse fields, we enhance the expressive ability of embeddings by using self-attention. The specific formula is shown in following formula, where $d$ is the dimension of $\mathbf{e}$ and $n$ is the number of fields.
\begin{small}
\begin{equation}
    \mathbb{X}(\mathbf{e_i}) = \sum_{j=1}^{n} {  (Softmax(\frac{\mathbf{e_i} \cdot \mathbf{e_j} ^T}{\sqrt{d}}) \cdot \mathbf{e_j}) \textbf{1}_{[i\neq j]} } 
\end{equation}
\end{small}
\vspace{-5pt}

%这其中的含义是，对于更近似的特征，其 miscalibration 的分布会相似，其Embedding $\mathbf{e_i}$ 与 被增强的  Embedding \mathbf{e_j} 的 内积会更大，因此对应的权重会更高。
That means: for more similar fields, their miscalibration distributions will be similar. If the semantics between two embeddings $\mathbf{e_i}$ and $\mathbf{e_j}$ is similar, then the corresponding weight is also relatively large.
\begin{small}
\begin{equation}
\label{eq-mlp-for-wei}
   \mathbf{\alpha}_i = Softmax(MLP(\mathbf{b_{\hat{p}_{uncalib}}}, \; \mathbf{e_{i}}, \; \mathbb{X}(\mathbf{e_i})))
\end{equation}
\end{small}

Next, we concatenate the original embedding $\mathbf{e_{i}}$, the enhanced output embedding $\mathbb{X}(\mathbf{e_i})$ , and the pCTR bucket embedding $\mathbf{b_{\hat{p}_{uncalib}}}$. Then, following the same process as in the shape allocation stage, we generate the attention for shape allocation $ \mathbf{\alpha}_i$ and perform the final fusion.

\subsubsection{Multi-Field Shape Ensemble} 

%在电商场景，我们需要校准的feature会有很多，而且不同的feature的校准值会产生冲突，举个例子，一个19~26岁女性对女鞋的需求上，在item类目-女鞋条件下，pctr在0.01~0.03区间被高估30\%，然而在user年龄段：19~26岁的条件下，pctr 在0.01~0.03区间被低估20%。因此我们不仅要对单个特征进行shape Calibration，还需要再全局角度对于不同的field 的shape Calibration的输出进行融合，从而在全局角度让校准的误差更小 

Different calibration values for different fields can conflict with each other. For example, there is a scenario where 19 to 26-year-old users demand for women's shoes. Under the item category field of "women's shoes", the non-calibrated model behaves 30\% over-estimation of pCTR in the 0.01 to 0.03 range. However, under the user age field of "19 to 26", there's a 20\% under-estimation of pCTR in the same 0.01 to 0.03 range. 

Consequently, we not only need to perform shape calibration on individual field but also need to globally harmonize the outputs of shape calibrators for different fields. This helps reduce calibration errors at a global level.
%我们使用Global Shape Attention 来融合不同Feature 得到的 输出结果。Global Shape Attention $\Psi$ 来自于 Global Shape Attention Generator 模块，$\Psi$ 的size 为 $1 \times n$ $\Psi_i \in [0,1]$, 并且 $\sum_{i=1}^{n} {\Psi_i} = 1$ 。如何生成 Global Shape Attention 将在 4.2.2 章节进行阐述.
%融合公式如下，一共有 $n$个特征,其中 \mathbb{S}(x)_i 为每一个 feature 的 shape Calibrator 输出。

We use Global Shape Attention to combine the output results obtained from different fields (Figure \ref{fig:enter-label-overall}a). Global Shape Attention, denoted as $\Psi$, is derived from the Global Shape Attention Generator module (details will be elaborated in section 4.2.2). The size of $\Psi$ is $1 \times n$, with $\Psi_i \in [0,1]$, and $ \sum_{i=1}^{n} {\Psi_i} = 1 $. The fusion formula is as follows, where there are a total of $n$ fields, and $\mathbb{S}_i(x)$ represents the output of the shape calibrator for $i$-th field.

% \vspace{-5pt}
\begin{small}
\begin{equation}
    \mathbb{S}(x) =\sum_{i=1}^{n} {\mathbb{S}_i(x) \cdot \Psi_i}
    \label{eq-s-multi}
\end{equation}
\end{small}

\vspace{-5pt}

We finally obtain the output of Global Shape Calibrator: $\mathbb{S}(x)$.

% \vspace{-10pt}
\subsection{Value Calibrator}
%%
%Value Calibrator的作用，对于每一个Feature Value，整体不会存在高估或者低估的情况。
% 1. 我们引入全部的特征，来进行 Value Calibration，使得整体的效果最优 4.2.1
% 2. 因为考虑全部特征，因此可以对于每一个Feature的Shape进行很好的选择，4.2.2

The goal of the Value Calibrator is to ensure that, for each sample $x$, there is no overall over- or under-estimation (value calibration).
We use all fields for Value Calibrator to achieve the best overall performance, as described in section 4.2.1. 
Considering all fields allows for the excellent allocation of shapes for each field, we use global shape attention depicted in section 4.2.2.

\subsubsection{Global Field Value Calibrator} Global Field Value Calibrator encompasses all the necessary information of the fields concerned. 
% 在这里我们参考了FAC的思路
We train a neural network to fix the field-level miscalibration or biases by utilizing all necessary features\cite{pan2020field}. These input fields are projected into fixed-size dense representations, which are fed into a neural network (the form of the neural network is not restricted, here we use MLP). Then we get the middle output $hiddenLayer$ and the final output $\mathbb{V}(x)$ for the Global Field Value Calibrator. 
%参考FAC%
Finally, by combining $\mathbb{V}(x)$ and $\mathbb{S}(x)$, we obtain the final calibrated output $\hat{p}_{calib} = \mathbb{S}(x) \cdot \mathbb{V}(x)$.
\begin{small}
\begin{equation}
    hiddenLayer = MLP_1(Concat(\mathbf{b_{\hat{p}_{uncalib}}},\mathbf{e}_1,\mathbf{e}_2,...,\mathbf{e}_n))
\end{equation}
\begin{equation}
    %\mathbb{V}(x) = MLP_1(MLP_2(Concat(\vec{bucket_{\hat{p}}},\mathbf{e_1},\mathbf{e_2},...,\mathbf{e_n})))
    \mathbb{V}(x) = MLP_2(hiddenLayer)
\end{equation}
\end{small}

\subsubsection{Global Shape Attention Generator}

% Global Feature Value Calibrator 除了能产出最终的输出值之外，还可以产出很有价值的中间数据。
% 在这里我们将中间层的节点作为输入，这部分能很好地学习全部特征信息，然后经过全连接网络，得到size 为 Feature Num 的输出值，经过softmax，得到 Global Shape Attention Score。在Shape Calibration 阶段，每一个Feature 产出的输出与其相乘求和，得到最终的 Shape Calibration Score。
% 在后面的消融实验中，我们也发现，对比后面接入MLP的方法，通过Global Shape Attention 的方法带来了明显的效果提升。
% 除了全部的样本特征值外，我们也把pctr 进行分桶后的特征加入 Global Shape Attention 的底层输入，这样会让 Global Feature Value Calibrator 以及 Global Shape Attention 在不同的pctr所有区间上，捕捉到差异化的信息，从而都能有更好的效果。

The Global Field Value Calibrator not only produces the final output but also yields valuable intermediate representation. In this context, we use the intermediate representation $hiddenLayer$ as input of Equation \ref{eq-psi-soft}, which can effectively learn information from all fields. After passing through a MLP and softmax layer, it produces Global Shape Attention $\Psi$ with a size equal to the number of fields $n$. In the Shape Calibration stage, the output produced by each field $Z_i$ is multiplied by its corresponding score $\Psi_i$ and summed to obtain the final Shape Calibration Score.
\begin{small}
\begin{equation}
    \Psi = Softmax(MLP_3(hiddenLayer))
\label{eq-psi-soft}
\end{equation}
\end{small}

% In subsequent ablation experiments, we also found that the Global Shape Attention operation has a significant performance improvement compared to directly connecting the MLP afterward.

\vspace{-10pt}
\subsection{Online Deployment}
%线上实验部署，内容包括：1）离线如何训练模型；2）考虑了哪些Field；3）离线方式提前算好所有Field的数据；4）线上如何拉到结果。最好画一个线上部署的流程图

\begin{figure}
\setlength{\abovecaptionskip}{0cm}
\setlength{\belowcaptionskip}{0cm}
    \centering
    \includegraphics[width=0.4\textwidth]{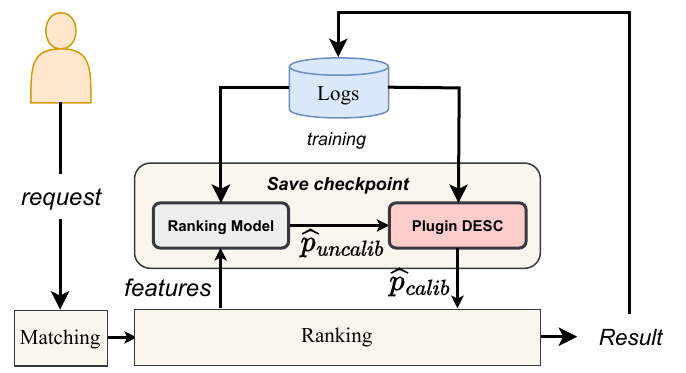}
    \caption{Real-Time calibration system used DESC method for CXR (CTR/CVR) task in our industrial advertising system.}
    \label{fig-online-dep}
    \vspace{-15pt}
\end{figure}

The overall framework of the online service using the DESC calibration method in our industrial advertising system is shown in Figure \ref{fig-online-dep}. When a real-time user request comes, all candidate items are recalled and predicted with pCTRs by a non-calibration model (ranking model). Then DESC model calibrates the pCTRs considering different field values, including user, item, and context features. In detail, the input $x$ of DESC has two parts: different field values and its non-calibrated score $\hat{p}_{uncalib}$. DESC uses a light neural network to output the calibrated score $\hat{p}_{calib}$, which is then used to sort the final ranks for all candidate items. We have integrated the DESC method as a plugin into the ranking model to reduce maintenance costs while only adding a little more time (about 2ms for one user's request) for online inference.

% \vspace{-10pt}
\section{EXPERIMENTS}
\subsection{Experimental Setup}
% 内容包括：1）介绍数据集（Train、Dev、Test数据量）；2）训练模型的超参数；3）评测指标（因为之前已经详细列举了指标的公式，所以这里只是描述一下要评测的指标）
% \subsubsection{Datasets.} To validate the effectiveness of our proposed DESC method, we conduct experiments on two public datasets (CTR prediction tasks) and one industrial dataset (CVR prediction task). The first public dataset is \textbf{Criteo} display advertising data \footnote{ \href{https://www.kaggle.com/c/criteo-display-ad-challenge}{https://www.kaggle.com/c/criteo-display-ad-challenge}}, which consists of 46 million samples over 10 days. We split the samples into 28 million samples as training set ($\mathcal{D}_{train}$), 9 million samples as validation set ($\mathcal{D}_{dev}$) and other 9 million samples as testing set ($\mathcal{D}_{test}$). Another public dataset is \textbf{AliCCP} (Alibaba Click and Conversion Prediction) \cite{ma2018entire} \footnote{ \href{https://tianchi.aliyun.com/datalab/dataSet.html?dataId=408}{https://tianchi.aliyun.com/datalab/dataSet.html?dataId=408}}, which contains 80 million samples. We split these samples with the proportion 2:1:1 to $\mathcal{D}_{train}$, $\mathcal{D}_{dev}$ and $\mathcal{D}_{test}$. 
% To test DESC method on CVR prediction task, we collect the conversion logs from the Shopee's online advertising system. It contains 100 million samples over 10 days, where the first 6 days for $\mathcal{D}_{train}$, the next 2 days for $\mathcal{D}_{dev}$ and the last 2 days for $\mathcal{D}_{test}$.

\subsubsection{Datasets.} To validate the effectiveness of our proposed DESC method, we conduct experiments on two public datasets (\textbf{AliCCP} and \textbf{CRITEO} for CTR prediction tasks) and one industrial dataset (\textbf{Shopee} for CVR prediction task). \textbf{CRITEO} display advertising data \footnote{ \href{https://www.kaggle.com/c/criteo-display-ad-challenge}{https://www.kaggle.com/c/criteo-display-ad-challenge}}, which consists of 46 million samples. We split the samples into 28 million samples as training set ($\mathcal{D}_{train}$), 9 million samples as validation set ($\mathcal{D}_{dev}$) and other 9 million samples as testing set ($\mathcal{D}_{test}$). \textbf{AliCCP} (Alibaba Click and Conversion Prediction) \cite{ma2018entire} \footnote{ \href{https://tianchi.aliyun.com/datalab/dataSet.html?dataId=408}{https://tianchi.aliyun.com/datalab/dataSet.html?dataId=408}} contains 80 million samples. We split these samples with the proportion 2:1:1 to $\mathcal{D}_{train}$, $\mathcal{D}_{dev}$ and $\mathcal{D}_{test}$. 
To test DESC method on CVR prediction task, we collect the conversion logs from the \textbf{Shopee}'s online advertising system. It contains 100 million samples, where the first 60 million for $\mathcal{D}_{train}$, the next 20 million for $\mathcal{D}_{dev}$ and the last 20 million for $\mathcal{D}_{test}$.
% over 10 days

\vspace{-5pt}
\subsubsection{Competing Methods.} Several representative calibration methods are used as competitors. We tested several competitive methods, including parametric methods, non-parametric methods, non-field-aware hybrid methods and field-aware hybrid methods. Parametric method contains Platt Scaling (PS) \cite{platt1999probabilistic}. Non-parametric method contains Histogram Binning (HB) \cite{zadrozny2001obtaining} and Istonic Regression (IR) \cite{zadrozny2002transforming}. Non-field-aware method contains Smooth Isotonic Regression (SIR) \cite{jiang2011smooth} and Ensemble Temperature Scaling (ETS) \cite{zhang2020mix}. Field-aware method contains FAC \cite{pan2020field}, Ada \cite{wei2022posterior} and MBCT \cite{huang2022mbct}.

We use DeepFM \cite{guo2017deepfm} to train the non-calibrated models with all fields in $\mathcal{D}_{train}$, and predict the non-calibrated scores in $\mathcal{D}_{dev}$ and $\mathcal{D}_{test}$. 
DeepFM consists of both the fully connected part and the FM part. We utilized the open-source DeepFM "from deepctr.models import DeepFM" and incorporated all features into DeepFM.
% DeepFM contains DNN Part and FM part and then sums these two part scores as the final non-calibrated score. % [TODO] Details of DeepFM structure.

All calibration methods are trained using samples of $\mathcal{D}_{dev}$ and are tested using samples of $\mathcal{D}_{test}$.
\vspace{-5pt}
\subsubsection{Parameter Configuration.} For all neural calibrators, including FAC, Ada and DESC, we use Adam as the optimizer with a learning rate of 1e-3 and batch size of 16,384. For the embedding size of field value $d$, both FAC and Ada are set as 256 while for DESC, it is 128 because it obtains a better result. For FAC, we set the number of bins to 100 while for Ada, the candidate set of bin numbers is set \{2, 4, 8\} following in \cite{wei2022posterior}. For DESC, the number of basis functions $m$ is 48 (the numbers of the three types of basis functions are both 16), and the numbers of fields $n$ are 26, 23 and 10 for Criteo, AliCCP and Industrial dataset, respectively. For MBCT, we set the same hyperparameters according to the paper.

\vspace{-5pt}
\subsubsection{Compared Metrics.}

Two commonly used metrics, F-RCE (Field RCE) and F-ECE (Field ECE) for each field $Z_i$ are used. To calculate the F-RCE of $i$-th field (shown in Equation \ref{eq-field-rce}), we use the testing dataset $\mathcal{D}_{test}$ (for simplicity, $\mathcal{D}$ refers to $\mathcal{D}_{test}$ here) consisting of $(x^j, y^j)$, where $x^j$ and $y^j$ mean the input features with different fields and clicked label of $j$-th sample. The subset of $\mathcal{D}^z$ has the samples with the same value $z$ for $i$-th field. F-RCE of $i$-th field evaluates the deviation level of each sample's calibrated probability $\hat{p}_{calib}^j$ considering $j$-th field. 

Another metric is the F-ECE of $i$-th field. As shown in Equation \ref{eq-field-ece}, it can be calculated for each subset $D^z$ of samples with the same value in $i$-th field. We can calculate the subset of F-ECE (shown in Equation \ref{eq-field-ece-M}) by partitioning predictions into $M$ equally-spaced bins and taking a weighted average of the difference between bins' accuracy and confidence. In detail, firstly we sort the samples by the non-calibrated scores $\hat{p}_{uncalib}^j$. Then all samples are grouped into $M$ interval bins. For all samples ($B_m$) in $m$-th bin, we calculate the accuracy and confidence, where the accuracy is the average of labels and the confidence is the average of calibrated scores. 

% \vspace{-4pt}
\begin{small}
\begin{equation}
    \label{eq-field-rce}
    F\text{-}RCE_{i} = \frac{1}{|\mathcal{D}|}\sum_{z \in Z_i}\frac{|\sum_{(x^j, y^j) \in \mathcal{D}^z}(y^j - \hat{p}_{calib}^j)|}{\frac{1}{|\mathcal{D}^z|} \sum_{(x^j, y^j) \in \mathcal{D}^z} y^j}
\end{equation}
% \vspace{-6pt}
\begin{equation}
    \label{eq-field-ece}
    F\text{-}ECE_{i}@M = \frac{1}{|\mathcal{D}|} \sum_{z \in Z_i} |\mathcal{D}^z| \cdot (F\text{-}ECE_{\mathcal{D}^z}@M)
\end{equation}
% \vspace{-6pt}
\begin{equation}
    \label{eq-field-ece-M}
    F\text{-}ECE_{\mathcal{D}^z}@M = \sum_{m=1}^{M} \frac{|B_m|}{|\mathcal{D}^z|}|acc(B_m) - conf(B_m)|
\end{equation}
% \vspace{-6pt}
\begin{equation}
    \label{eq-field-ece-acc-conf}
    acc(B_m) = \frac{1}{|B_m|} \sum_{j \in B_m} y^j, \; conf(B_m) = \frac{1}{|B_m|} \sum_{j \in B_m} \hat{p}_{calib}^j
\end{equation}
\end{small}

As explained that the traditional field-aware calibration methods, such as FAC \cite{pan2020field} and Ada \cite{wei2022posterior}, only consider one field to calibrate, these methods do not take into account the influence of other fields on the calibration results. We found that the calibration results from these models only training on one field perform poorly on other fields. So we use the Multi-Field RCE (MF-RCE) and Multi-Field ECE (MF-ECE) on all fields to compare different methods as follows:
\vspace{-5pt}
\begin{small}
\begin{equation}
    MF\text{-}RCE = \frac{1}{n} \sum_i F\text{-}RCE_{i} 
\end{equation}

\begin{equation}
    MF\text{-}ECE@M = \frac{1}{n} \sum_i F\text{-}ECE_{i}@M ,
\end{equation}
\end{small}
where $n$ is the number of fields.

%The Field-RCE and Field-ECE only reflect the calibration error on a certain group of partitions according to the specific field. Ideally, a well-performing calibrator should has a calibration error of nearly zero under any partition. %Then, the multi-view calibration error (MVCE) proposed in MBCT \cite{huang2022mbct} is used to compare different calibrators. MVCE metric can be calculated in Equation \ref{eq-mvce}, where $r$ is the number of partitions, $t_i$ is the number of bins under $i$-th partition, $D_{i,j}$ refers to the samples of $j$-th bin under $i$-th partition, $p$ is the hyper-parameter.
% \begin{equation}
%     \label{eq-mvce}
%     MVCE = (\frac{1}{r} \sum_{i=1}^r (\frac{1}{t_i} \sum_{j=1}^{t_i} |acc(D_{i,j}) - conf(D_{i,j})|)^p)^{\frac{1}{p}}
% \end{equation}

We use F-RCE and F-ECE as the primary empirical metrics to compare the calibration errors of different calibration methods on one field and MF-RCE and MF-ECE to test on all fields. Besides, we also report the overall ranking performance by using AUC and Log-loss metrics. Since PCOC (Predicted Click Over Click) \cite{he2014practical, graepel2010web} and ECE (Expected Calibration Error) \cite{huang2022mbct} indicators can be achieved well for all competing methods, and F-ECE and MF-ECE, as compared to the traditional ECE, can provide a finer-grained representation of calibration error in the context of field-aware problems, PCOC and ECE are not listed in the main results.

\vspace{-5pt}
\subsection{Main Results}
% 主要实验结果，Table1大表展示各方法在各数据、各指标的效果（均是离线实验结果）

\begin{table*}
\setlength{\abovecaptionskip}{0cm}
\setlength{\belowcaptionskip}{0cm}
    \center
    \setlength\tabcolsep{1.0pt}
  \caption{Results of different methods for calibrating CTR predictive models on public and industrial datasets for one field.}  %\tnote{1}
  \label{tab-main-one}
  \begin{threeparttable}
  \small
  %\resizebox{\linewidth}{!}
  \begin{tabular}{c|c|cccc|cccc|cccc}
  \toprule[1.2pt]
    % \hline
    \multirow{2}*{Type} & \multirow{2}*{Method} & \multicolumn{4}{c}{Criteo} & \multicolumn{4}{c}{AliCCP} & \multicolumn{4}{c}{Industrial Data} \\
    \cline{3-14}
    ~ & ~ & AUC $\uparrow$ & Log-loss $\downarrow$ & F-RCE $\downarrow$ & F-ECE@3 $\downarrow$ & AUC $\uparrow$ & Log-loss $\downarrow$ & F-RCE $\downarrow$ & F-ECE@3 $\downarrow$ & AUC $\uparrow$ & Log-loss $\downarrow$ & F-RCE $\downarrow$ & F-ECE@3 $\downarrow$ \\
    \hline
    No Calib. & N/A & 0.7949 & 0.4559 & 0.0426 & 0.0119 & 0.6130 & 0.1611 & 0.0876 & 0.0135 & 0.8613 & 0.5212 & 0.0743 & 0.0253 \\
    \hline
    \multirow{2}*{Non-Param} & HB & 0.7944 & 0.4562 & 0.0328 & 0.0112 & 0.6128 & 0.1621 & 0.0852 & 0.0130 & 0.8610 & 0.5218 & 0.0669 & 0.0204 \\
    ~ & IR & 0.7949 & 0.4559 & 0.0320 & 0.0110 & 0.6130 & 0.1611 & 0.0843 & 0.0132 & 0.8613 & 0.5212 & 0.0665 & 0.0202 \\
    \hline
    Param & PS & 0.7949 & 0.4559 & 0.0301 & 0.0105 & 0.6130 & 0.1611 & 0.0841 & 0.0128 & 0.8613 & 0.5212 & 0.0660 & 0.0201 \\
    \hline
    \multirow{2}*{\makecell{
    Non-Field- \\
    Aware}} & SIR & 0.7949 & 0.4559 & 0.0300 & 0.0097 & 0.6130 & 0.1611 & 0.0836 & 0.0127 & 0.8613 & 0.5211 & 0.0660 & 0.0201 \\
    ~ & ETS & 0.7949 & 0.4559 & 0.0293 & 0.0095 & 0.6130 & 0.1611 & 0.0837 & 0.0127 & 0.8613 & 0.5212 & 0.0661 & 0.0199 \\
    \hline
    \multirow{4}*{\makecell{Field- \\ Aware}} & FAC & 0.7991 & 0.4523 & 0.0265 & 0.0087 & 0.6751 & 0.1571 & 0.0829 & 0.0123 & 0.8634 & 0.5203 & 0.0402 & 0.0163 \\
    ~ & Ada & 0.7992 & \textbf{0.4521} & 0.0273 & 0.0089 & 0.6752 & 0.1568 & 0.0832 & 0.0125 & 0.8639 & 0.5201 & 0.0413 & 0.0170 \\
    ~ & MBCT & 0.7951 & 0.4550 & 0.0248 & 0.0095 & 0.6742 & 0.1611 & 0.0828 & 0.0128 & 0.8621 & 0.5209 & 0.0409 & 0.0175 \\
    ~ & DESC & \textbf{0.7992} & 0.4522 & \textbf{0.0203} & \textbf{0.0080} & \textbf{0.6774} & \textbf{0.1566} & \textbf{0.0821} & \textbf{0.0120} & \textbf{0.8641} & \textbf{0.5192} & \textbf{0.0307} & \textbf{0.0132}\\
  \bottomrule[1.2pt]
\end{tabular}
% \begin{tablenotes}
% \footnotesize
% % \item[1] We use [[[xxx, yyy and zzz]]] fields while for Criteo, AliCCP and Industrial dataset, respectively.
% % \item[2]"FR" refers to the Field-RCE.
% % \item[3]"FE" refers to the Field-ECE.
% % \item[4]"M" refers to the Multi-Field-ECE.
% \end{tablenotes}
\end{threeparttable}
\end{table*}

% \vspace{-5pt}
\begin{table*}
\setlength{\abovecaptionskip}{0cm}
\setlength{\belowcaptionskip}{0cm}
    \center
  \caption{Results of different methods for calibrating CTR predictive models on public and industrial datasets for all fields.}  %\tnote{1}
  \label{tab-main}
  \begin{threeparttable}
  \small
  %\resizebox{\linewidth}{!}
  \begin{tabular}{c|c|cc|cc|cc}
    \toprule[1.2pt]
    \multirow{2}*{Type} & \multirow{2}*{Method} & \multicolumn{2}{c}{Criteo} & \multicolumn{2}{c}{AliCCP} & \multicolumn{2}{c}{Industrial Data} \\
    \cline{3-8}
    ~ & ~ & MF-RCE $\downarrow$ & MF-ECE@3 $\downarrow$ & MF-RCE $\downarrow$ & MF-ECE@3 $\downarrow$ & MF-RCE $\downarrow$ & MF-ECE@3 $\downarrow$ \\
    \hline
    No Calib. & N/A & 0.0702 & 0.0142 & 0.1492 & 0.0143 & 0.0793 & 0.0370 \\
    \hline
    \multirow{2}*{Non-Param} & HB & 0.0692 & 0.0134 & 0.1430 & 0.0135 & 0.0779 & 0.0250 \\
    ~ & IR & 0.0688 & 0.0131 & 0.1427 & 0.0131 & 0.0766 & 0.0249 \\
    \hline
    Param & PS & 0.0685 & 0.0130 & 0.1420 & 0.0129 & 0.0765 & 0.0248 \\
    \hline
    \multirow{2}*{\makecell{
    Non-Field- \\
    Aware}} & SIR & 0.0686 & 0.0131 & 0.1422 & 0.0130 & 0.0764 & 0.0247 \\
    ~ & ETS & 0.0683 & 0.0127 & 0.1470 & 0.0129 & 0.0761 & 0.0248 \\
    \hline
    \multirow{4}*{\makecell{Field- \\ Aware}} & FAC & 0.0517 & 0.0101 & 0.1321 & 0.0120 & 0.0412 & 0.0183 \\
    ~ & Ada & 0.0501 & 0.0102 & 0.1338 & 0.0122 & 0.0415 & 0.0180 \\
    ~ & MBCT & 0.0660 & 0.0104 & 0.1357 & 0.0125 & 0.0430 & 0.0192 \\
    ~ & DESC & \textbf{0.0489} & \textbf{0.0099} & \textbf{0.1264} & \textbf{0.0115} & \textbf{0.03606} & \textbf{0.0132} \\
  \bottomrule[1.2pt]
\end{tabular}
% \begin{tablenotes}
% \footnotesize
% \end{tablenotes}
\end{threeparttable}
\end{table*}

\subsubsection{Results on One Field.} As other hybrid field-aware based methods only consider one field to calibrate, we compare our proposed method in one field in the same way. We select the field of "C2" (577 unique values) in Criteo, "853" (39,979 unique values) in AliCCP and "item ID" (5 million unique values) in Industrial data to perform our experiments. As shown in Table \ref{tab-main-one}, the AUCs of all calibrators are higher than the uncalibrator while all other calibration-related metrics (F-ECE, F-RCE) of calibrators are smaller than the uncalibrator, which shows that the calibration methods can improve the accuracy of pCTR while maintaining a certain increase in ranking performance. Specially, by fusing multiple fields in shape allocation and shape augmentation, our proposed DESC can achieve much smaller calibration errors compared to other competitors even in one field evaluation. It's worth explaining that DESC performs much better than other competitors in industrial data than public data. The reason is that DESC behaves better in calibration performance in this scene where the data sparsity is more serious in practical industrial data.  

% \begin{table*}
%     \center
%   \caption{Results of different methods for calibrating CTR predictive models on public and industrial datasets for one field\tnote{1}.}
%   \label{tab-main}
%   \begin{threeparttable}
%   \small
%   %\resizebox{\linewidth}{!}
%   \begin{tabular}{c|c|ccccc|ccccc|ccccc}
%     \hline
%     \multirow{2}*{Type} & \multirow{2}*{Approach} & \multicolumn{5}{c}{Criteo} & \multicolumn{5}{c}{AliCCP} & \multicolumn{5}{c}{Industrial Data} \\
%     \cline{3-17}
%     ~ & ~ & AUC & FR\tnote{2} & FE\tnote{3}@3 & FE@5 & M\tnote{4} & AUC & FR & FE@3 & FE@5 & M & AUC & FR & FE@3 & FE@5 & M \\
%     \hline
%     No Calib. & N/A & - & - & - & - & - & - & - & - & - & - & - & - & - & - & - \\
%     \hline
%     \multirow{2}*{\makecell{Hybrid \\
%     Non-field-aware}} & SIR & - & - & - & - & - & - & - & - & - & - & - & - & - & - & - \\
%     ~ & ETS & - & - & - & - & - & - & - & - & - & - & - & - & - & - & - \\
%     \hline
%     \multirow{4}*{\makecell{Hybrid field-aware}} & FAC & - & - & - & - & - & - & - & - & - & - & - & - & - & - & - \\
%     ~ & Ada & - & - & - & - & - & - & - & - & - & - & - & - & - & - & - \\
%     ~ & MBCT & - & - & - & - & - & - & - & - & - & - & - & - & - & - & - \\
%     ~ & DESC & - & - & - & - & - & - & - & - & - & - & - & - & - & - & - \\
%   \bottomrule
% \end{tabular}

\subsubsection{Results on All Fields.}
As DESC can consider the effects of different fields on calibration, we need to evaluate its performance on all fields. As shown in Table \ref{tab-main}, the conclusion is consistent with the smallest calibration errors on all three datasets.

\vspace{-5pt}
\subsection{In-Depth Analysis}
\subsubsection{Model Structure Ablation Experiment for DESC}
We perform some model structure ablation experiments to validate the significance and effectiveness.  

{\bfseries(1) Ablation Experiment of Shape Calibrator.} To evaluate the effectiveness of the Shape Calibrator and Value Calibrator, we delete the corresponding module and left the other parts unchanged. Firstly, we delete the Shape Calibrator sub-module, the AUC reduces to 0.6602 while the metrics of MF-RCE and MF-ECE@3 on all fields are slightly larger than DESC with 0.1434 and 0.0120 (first line on Table. \ref{tab-pm-exp}).

% 为了证明 Shape 的重要性
{\bfseries(2) Ablation Experiment of Value Calibrator.} Secondly, we delete another sub-module, the Value Calibrator. The phenomenon is almost the same, with 0.6761 of AUC, and 0.1398 and 0.0119 of MF-RCE and MF-ECE@3 on all fields (second line on Table. \ref{tab-pm-exp}).

% 为了证明 Value 的重要性
{\bfseries(3) Ablation Experiment of Multi-Field Shape Ensemble.} In METHODS section, we have emphasized that shape calibration needs to consider multiple fields because different fields and values can have numerous different shapes. Then, we only use mean-pooling to compute the final predicted result considering different fields ($\Psi_i$ is all the same with $1/n$ in Equation \ref{eq-s-multi}). The results are still inferior as shown in the third line in Table. \ref{tab-pm-exp}.

% 为了证明 不同 Field 融合，用 Attention Ensemble 的必要性
{\bfseries(4) Ablation Experiment of pCTR Bucket Feature.} The pCTR bucket feature is concatenated with the field embedding to calculate the shape allocation weights in DESC (Equation \ref{eq-mlp-for-wei}). Then we delete the pCTR bucket feature (fourth line on Table. \ref{tab-pm-exp}) to verify that different pCTR values can also have effects on the calibration shape.

% 为了证明 pCTR 在选择Shape时，倾向性的重要性
{\bfseries(5) Ablation Experiment of Embedding Augmentation.} To enhance the expressive ability of embeddings, the self-attention mechanism is used. Then we delete this part by using the original Equation \ref{eq-mlp-ori-wei} rather than Equation \ref{eq-mlp-for-wei}. The fifth line on Table. \ref{tab-pm-exp} shows that the embedding augmentation mechanism can enhance the representational power of embedding. 

% 为了证明 Embedding Augment 的确可以缓解系数 特征的校准效果

% 消融实验：1）w/o Sigmoid曲线；2）Sigmoid曲线不与Field进行Attention；3）Field Attention来计算Bias；【可做可不做 4）Sigmoid曲线的数量分析；】

% \begin{table}
% \renewcommand\arraystretch{1.0}
%   \caption{Ablation study of DESC on AliCCP.}
%   \label{tab-pm-exp}
%   \begin{threeparttable}
%   \small
%   \begin{tabular}{c|ccccc}
%     \hline
%     Type & AUC & FR & FE@3 & FE@5 & M \\
%     \hline
%     w/o Shape Calibrator & - & - & - & - & - \\
%     w/o Value Calibrator & - & - & - & - & - \\
%     w/o Multi-Field Shape Ensemble & - & - & - & - & - \\
%     w/o pCTR Bucket Feature & - & - & - & - & - \\
%     w/o Embedding Augment & - & - & - & - & - \\
%     DESC & - & - & - & - & - \\
%   \bottomrule
% \end{tabular}
% \begin{tablenotes}
% \footnotesize
% % \item[1] "SC" refers to Shape Calibrator.
% % \item[2] "VC" refers to Value Calibrator.
% % \item[3] "MSE" refers to Multi-Field Shape Ensemble.
% % \item[4] "PBF" refers to pCTR Bucket Feature.
% % \item[5] "EA" refers to Embedding Augment.
% \end{tablenotes}
% \end{threeparttable}
% \end{table}

\begin{table}
\setlength{\abovecaptionskip}{0cm}
\setlength{\belowcaptionskip}{0cm}
\setlength\tabcolsep{1.0pt}
\renewcommand\arraystretch{1.0}
  \caption{Ablation study of DESC on AliCCP.}
  \label{tab-pm-exp}
  \begin{threeparttable}
  \small
  \begin{tabular}{c|cccc}
    \toprule[1.2pt]
    Type & AUC $\uparrow$ & Log-loss $\downarrow$ & MF-RCE $\downarrow$ & MF-ECE@3 $\downarrow$ \\
    \hline
    w/o Shape Calibrator & 0.6602 & 0.1581 & 0.1434 & 0.0120 \\
    w/o Value Calibrator & 0.6761 & 0.1567 & 0.1398 & 0.0119 \\
    w/o Multi-Field Shape Ensemble & 0.6755 & 0.1568 & 0.1324 & 0.0116 \\
    w/o pCTR Bucket Feature & 0.6764 & 0.1567 & 0.1391 & 0.0120 \\
    w/o Embedding Augment & 0.6764 & 0.1567 & 0.1398 & 0.0119 \\
    \hline
    DESC & \textbf{0.6774} & \textbf{0.1566} & \textbf{0.1264} & \textbf{0.0115} \\
  \bottomrule[1.2pt]
\end{tabular}
% \begin{tablenotes}
% \footnotesize
% % \item[1] "SC" refers to Shape Calibrator.
% % \item[2] "VC" refers to Value Calibrator.
% % \item[3] "MFSE" refers to Multi-Field Shape Ensemble.
% % \item[4] "PBF" refers to pCTR Bucket Feature.
% % \item[5] "EA" refers to Embedding Augment.
% \end{tablenotes}
\end{threeparttable}
\vspace{-10pt}
\end{table}

\vspace{-5pt}
\subsubsection{DESC exhibits a stronger data utilization capability.} In this part, we analyze the data utilization capabilities of different calibrators from both a global perspective (overall down sampling) and an individual perspective (field value sample quantity).

\begin{figure}
\setlength{\abovecaptionskip}{0cm}
\setlength{\belowcaptionskip}{0cm}
    \centering
    \includegraphics[width=0.7\linewidth]{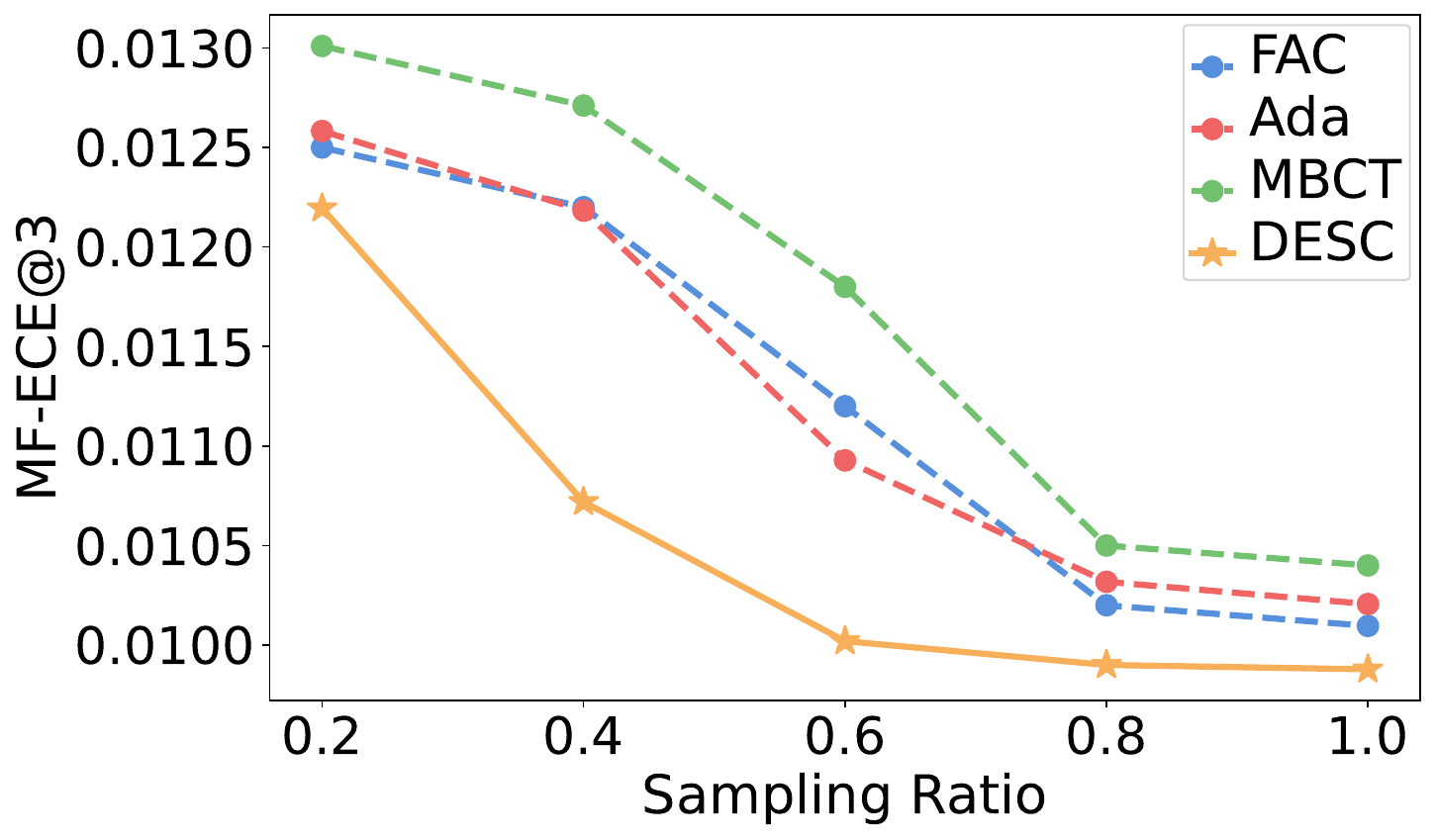}
    \caption{Down sampling analysis in CRITEO.}
    \label{fig-data-sparse}
    \vspace{-10pt}
\end{figure}

{\bfseries(1) Overall Down Sampling }
% 为了验证
We further conducted experiments to compare the calibration performance of four different hybrid field-aware methods (FAC, Ada, MBCT and DESC) across various sampling ratios. As shown in Figure \ref{fig-data-sparse}, DESC exhibits a distinct lower MF-ECE in the context of varying sampling ratios. In a global perspective, this demonstrates that DESC exhibits a stronger data utilization capability.

\begin{figure}
\setlength{\abovecaptionskip}{0.0cm}
\setlength{\belowcaptionskip}{-0.5cm}
    \centering
    \includegraphics[width=0.6\linewidth]{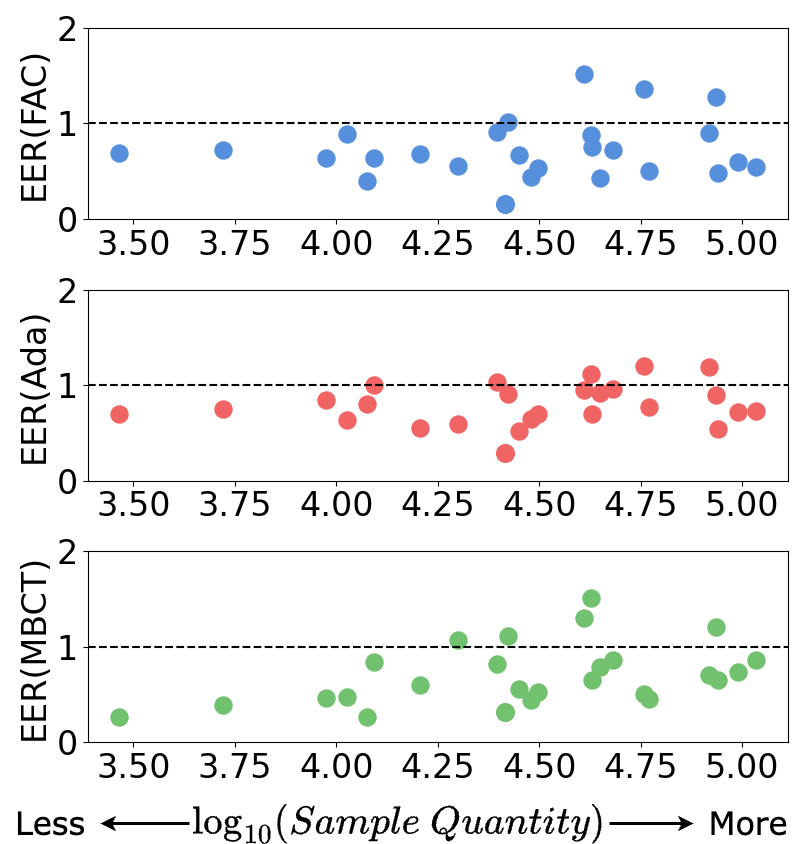}
    \caption{Performance differences for the sample quantity of the field value.}
    \label{fig:perf-fv}
    % \vspace{-10pt}
\end{figure}

{\bfseries(2) Field Value Sample Quantity.} We analysis the performance differences among MBCT, FAC, Ada and DESC at sample quantity aspect. We set the quantity of samples with a specific field value (for the convenience of display, we use $log_{10}(x)$) as the x-axis and $EER_{Calib}$ ($\mathbf{E}$xcept Calibration $\mathbf{E}$rror $\mathbf{R}$atio) as the y-axis (the smaller $EER_{Calib}$ is, the better DESC is compared with other calibrators), as shown in Figure \ref{fig:perf-fv}.
\begin{small}
\begin{equation}
\setlength{\abovedisplayskip}{2pt}
\setlength{\belowdisplayskip}{10pt}
    EER_{Calib} = \frac{ECE_{DESC}}{ECE_{Calib}}, Calib \in \{MBCT, FAC, Ada\}
\end{equation}
\end{small}
\vspace{-10pt}

ECE (Expected Calibration Error) \cite{huang2022mbct} is used as the calibration error to explain the results. The ECE value ranges from $(0, +\infty]$. The higher ECE value indicates higher calibration error and worse performance.
We can find that for the samples with small sized of number, $EER$ is less than 1, which means, from an individual perspective, that DESC possesses a stronger data utilization capability.
\vspace{-10pt}
\subsubsection{DESC Can Adapt to Complex Miscalibration Shapes.} %Estimated Bias Distributions
We analysis the performance differences among MBCT, FAC, Ada and DESC at shape complexity.

From the perspective of pCTR, we divide the estimated pCTR values into bins either with equal frequency or equal interval, and each bin can calculate the PCOC (Predicted Click Over Click) \cite{he2014practical, graepel2010web} to describe the over- and under-estimation within that bin.

% 对于一些情况，每个bin的高低估情况是接近的，有些情况每个bin的高低估情况是差距较大的，如Fig5。我们定义MisCalibration Complexity 指标来刻画不同pCTR分bin的高低估情况是差距，MisCalibration Complexity 越高，分bin的高低估情况是差距越大。在校准器的角度，MisCalibration Complexity 越高，校准的难度也越高，一方面对于 shape 准度要求越高，一方面过度依赖于后验数据使得校准器的泛化能力受损。

For some cases, the over- and under-estimation within each bin are close, while in other cases, there is a significant difference in over- and under-estimation for each bin. We define the "Miscalibration Complexity" metric to characterize the disparity in over- and under-estimation across different pCTR bins. We selected all field values for a subset of fields. For field $z$ with a value $v$, the samples are divided into $Q$ ($Q>1$) bins. The miscalibration complexity is denoted as $\mathbb{D}_{Z,v}$ as follows:
\begin{small}
\vspace{-5pt}
\begin{equation}
    \mathbb{D}_{Z,v} = \frac{1}{Q-1} \sum_{bin=1}^{Q-1} {\lvert PCOC_{bin+1} - PCOC_{bin} \rvert} .
\end{equation}
\end{small}
\vspace{-5pt}

In general, the higher the miscalibration complexity, the higher the disparity in over- and under-estimation within the bins. For a calibrator, a higher miscalibration complexity indicates a greater calibration challenge. On one hand, it requires higher precision in shaping the accuracy of the calibration, and on the other hand, an excessive reliance on posterior data can compromise the generalization capability of the calibrator.

We still use the ECE metric to characterize the calibration error of different calibrators under the specific field value. We compare MBCT, FAC, Ada, and DESC and obtain the $EER_{Calib}$.

\begin{figure}
\setlength{\abovecaptionskip}{0cm}
\setlength{\belowcaptionskip}{0cm}
    \centering
    \includegraphics[width=0.6\linewidth]{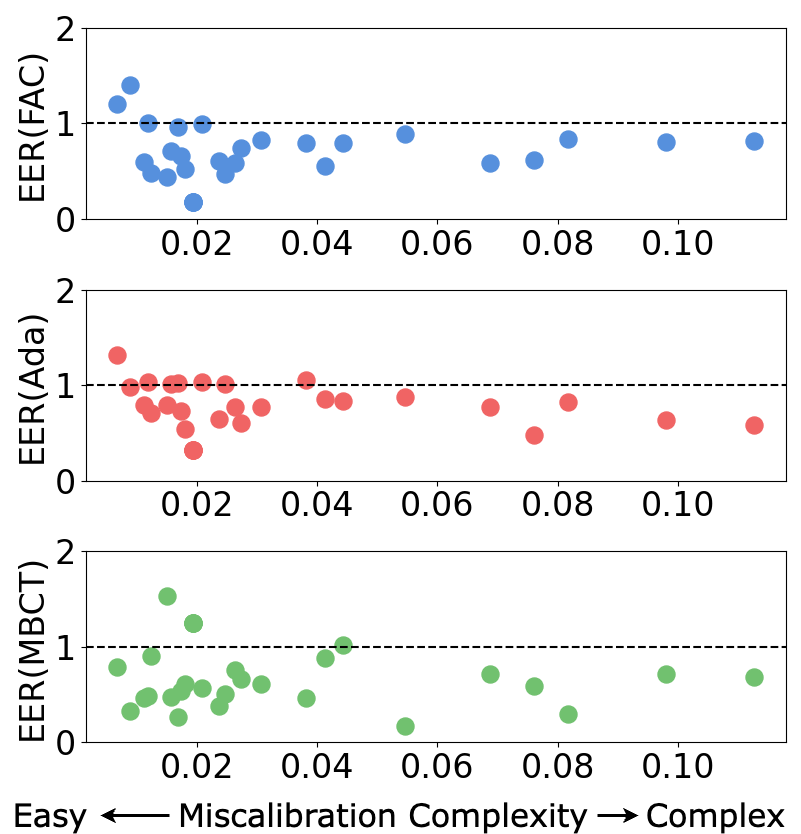}
    \caption{Performance differences for miscalibration complexity aspect.}
    \label{fig:perf-dif-asp}
    \vspace{-5pt}
\end{figure}
 
For each calibrator and each field value, there is a corresponding $\mathbb{D}_{Z,v}$ value as the x-axis and $EER_{Calib}$ as the y-axis. As shown in Figure \ref{fig:perf-dif-asp}, we can observe that in the high $\mathbb{D}_{Z,v}$ range, all $EER_{Calib}$ values are less than 1, which indicates that DESC always performs better in conditions of complex shapes compared to other methods.

\vspace{-8pt}
\subsection{Online Results}
To prove the effectiveness of the proposed method, we further conduct real-world online experiments. We feed item ID with 5 million unique values, category ID with 5,000 unique values, bidding type, time, raw predict score and other features into DESC. Then the model is deployed for online pCVR calibration on Shopee's advertising system. For the online A/B test, we build two experimental buckets, each with ten percent of the traffic. One bucket was configured with SIR \cite{jiang2011smooth} as the control group, while the other was configured with DESC as the experimental group. We use the two significant important business metrics CVR (Conversion Rate) and GMV (Gross Merchandise Volume) to evaluate our calibration effectiveness. The online results over a 7-day period indicated that DESC brings +2.5\% on CVR and +4.0\% on GMV compared to SIR. These improvements verify the effectiveness and business values of DESC in practical advertising system.

\vspace{-8pt}
\section{DISCUSSION}
We propose a new calibration approach named DESC. Firstly, we put forward multiple types of functions (\emph{e.g.}, power function, scaling function, logarithmic function) as the basis functions. Secondly, we allocate the appropriate basis functions to combine the shape function given the specific field and specific value. Finally, all fields are concatenated and used to calculate the weight for the calibration value from each field, which can further improve the accuracy of pCTR on multiple fields. Both offline and online experiments verify that DESC achieves significant improvement. In future work, we should explore the performance of our proposed method on other machine learning tasks except CXR task. Also, we need to study the reasons of miscalibration error in these machine learning tasks, especially for CXR task, and find new technology (\emph{e.g.} data augment, multi-tasking learning) to further improve the performance.
% 第一段：总结一下本文的创新点，离线和在线分析有多少的收益
% 第二段：对于其它方面的讨论，包括：1）本文有待改进，提高的点；2）本文方法应用于非CXR任务上的讨论；

\bibliographystyle{ACM-Reference-Format}
\balance
\bibliography{sample}
\end{document}